\documentclass[runningheads]{llncs}

\usepackage[T1]{fontenc}
\usepackage{CJKutf8}
\usepackage{graphicx}
\usepackage{booktabs}
\usepackage{array}
\usepackage{multirow}

\begin{document}

\title{X-MADAM-RAG: Diagnosing and Handling
Chinese-English Evidence Conflict in Retrieval-Augmented Generation}

\titlerunning{Diagnosing Chinese-English Evidence Conflict in RAG}

\author{Yongqi Kang\inst{} \and
Yu Fu\inst{} \and Yong Zhao\inst{*}}

\authorrunning{F. Author et al.}

\institute{Sichuan University, Chengdu, China\\
\email{2023141520239@stu.scu.edu.cn} \\
\email{fuyu05@stu.scu.edu.cn} \\
\email{yong.zhao@scupi.cn}}

\maketitle

\begin{abstract}
Retrieval-augmented generation (RAG) systems may receive evidence that is not
merely noisy but mutually contradictory. This issue becomes particularly salient
in multilingual settings, where retrieved Chinese and English evidence may
support incompatible answer candidates. We study this problem through
\textbf{X-RAMDocs-ZHEN}, a controlled Chinese-English benchmark derived from
RAMDocs for diagnosing evidence conflict in RAG. The benchmark contains 300
examples across six balanced conditions, including monolingual support,
bilingual agreement, reversed conflict directions, and conflict with optional
noise. We further examine \textbf{X-MADAM-RAG}, an interpretable pipeline that
decomposes evidence handling into per-document candidate extraction,
visible-evidence repair, deterministic candidate grouping, and conflict-aware
aggregation. On the original controlled benchmark with Qwen2.5-7B-Instruct,
X-MADAM-RAG achieves 0.9667 strict accuracy and 0.9767 conflict-aware success,
outperforming an evidence-normalized single-call baseline. However, a zero-call
rule-only extractor reaches 1.0000 on the same benchmark, revealing strong
template regularity. To probe this limitation, we construct a deterministic
naturalized stress test that removes explicit answer templates while preserving
candidate strings. On its 100-sample subset, rule-only extraction falls to
0.0000, but X-MADAM-RAG also drops to 0.3000 strict accuracy, below both naive
and evidence-normalized baselines. A privileged oracle remains perfect,
indicating that document-level extraction is the main bottleneck. These findings
position X-RAMDocs-ZHEN and X-MADAM-RAG as diagnostic tools for controlled
evidence conflict rather than as evidence of general hallucination detection or
robustness to natural retrieval.

\end{abstract}
\section{Introduction}

Retrieval-augmented generation has become a widely used approach for improving
the factual grounding of large language models. By conditioning generation on
external evidence, RAG systems aim to reduce reliance on parametric memory and
produce responses that are better supported by retrieved documents. However,
retrieved evidence is not always consistent. In realistic multilingual scenarios,
a system may retrieve Chinese and English sources that support different answers
to the same question---for example, a Chinese snippet may support one date while
an English snippet supports another, forcing the system to preserve disagreement
rather than silently select one candidate.

Most multilingual RAG work focuses on whether cross-lingual evidence can help
answer questions. We focus on a complementary diagnostic question: how do RAG
systems behave when evidence in different languages supports conflicting answers?
We are particularly interested in whether a system can preserve
evidence-supported candidates, recognize conflict, and avoid unsupported
candidate generation.

To study this, we construct \textbf{X-RAMDocs-ZHEN v0.1}, a controlled
Chinese-English benchmark with 300 examples across six balanced evidence
conditions. We also study \textbf{X-MADAM-RAG}, an interpretable pipeline that
extracts candidates per document, applies label-blind repair, groups candidates
deterministically, and aggregates the extracted candidates in a conflict-aware
manner. On the original benchmark,
X-MADAM-RAG substantially outperforms single-call LLM baselines, but a rule-only
extractor also solves it perfectly. A naturalized stress test removes explicit
answer templates: rule-only extraction collapses, but X-MADAM-RAG also degrades
below single-call baselines while a privileged oracle remains perfect,
identifying extraction as the primary bottleneck.

Our contributions are threefold:
\begin{enumerate}
\item We introduce \textbf{X-RAMDocs-ZHEN v0.1}, a leakage-aware controlled
Chinese-English evidence-conflict benchmark with six balanced conditions.
\item We present \textbf{X-MADAM-RAG}, an auditable pipeline with per-document
extraction, visible-evidence repair, deterministic grouping, and conflict-aware
aggregation.
\item We provide diagnostic analyses showing original-template success, ablation
contributions, and naturalized extraction fragility---identifying document-level
extraction robustness as the central open problem.
\end{enumerate}

We do not claim that the benchmark is human-validated, that X-MADAM-RAG solves
hallucination detection, or that the results generalize to natural multilingual
retrieval.

\section{Related Work}

Prior work on RAG faithfulness and hallucination diagnostics evaluates whether
generated outputs are supported by retrieved context. FaithEval examines model
faithfulness under parametric-contextual conflict~\cite{faithfuleval_2025},
FaithfulRAG models fact-level conflict~\cite{faithfulrag_2025}, RAGChecker
provides fine-grained RAG diagnostics~\cite{ragchecker_2024_or_2025}, and
RefChecker targets knowledge-centric hallucination~\cite{refchecker_2024_or_2025}.
Multilingual hallucination has also been studied in shared tasks such as
Mu-SHROOM~\cite{vazquez2025semeval}. Our work is narrower: we diagnose
answer-candidate behavior under controlled Chinese-English evidence conflict
using automatic lexical metrics rather than open-ended semantic faithfulness.

Knowledge conflict arises when retrieved evidence contains incompatible claims
or contradicts parametric knowledge~\cite{knowledge_conflict_survey_2024,cai2025practices,cai2024bringing,fu2025reasoning}.
RAMDocs and MADAM-RAG directly motivate our benchmark design by studying RAG
with conflicting evidence~\cite{wang2025retrieval}. We extend this setting
with a cross-lingual dimension: Chinese and English evidence are controlled to
agree, disagree, or reverse which language carries the reference candidate.
Decomposition-based approaches such as MADAM-RAG and FaithfulRAG process evidence
in smaller units before aggregation; X-MADAM-RAG follows this line but uses
deterministic candidate grouping after document-level extraction, making each
stage separately inspectable.

Multilingual and cross-lingual RAG systems retrieve and generate across
languages. XRAG studies cross-lingual retrieval-augmented
generation~\cite{xrag_2025}, and multilingual RAG work has examined
knowledge-intensive question answering across
languages~\cite{ranaldi2025multilingual}. Our work complements these by
focusing specifically on answer-level disagreement between Chinese and English
evidence, with controlled conditions that isolate conflict direction from
retrieval noise.

\section{X-RAMDocs-ZHEN Benchmark}

\subsection{Task and Dataset}

Given a question and a set of Chinese and/or English evidence snippets, the
system must produce an answer or a conflict-aware response using only the visible
evidence. Evidence may support a single candidate, multiple matching candidates,
or incompatible candidates across languages. A desirable system should preserve
evidence-supported candidates, recognize disagreement, avoid unsupported
candidates, and provide traceable evidence use. The task is diagnostic: evidence
conditions are controlled and private metadata is used only for construction and
evaluation.

\textbf{X-RAMDocs-ZHEN v0.1} is derived from RAMDocs and contains 300 samples
across 534 evidence documents, split evenly between Chinese and English.
Questions are bilingual (175 Chinese, 125 English). Conflicts cover entities
(144), dates (101), and numbers (55). The benchmark uses 54 unique source
records from 68 usable RAMDocs records, with 100\% provenance coverage and zero
automatic severe errors or warnings. No completed human audit is available.

\subsection{Evidence Conditions}

The benchmark contains six balanced evidence conditions (Table~\ref{tab:conditions}).

\begin{table}[t]
\caption{The six balanced evidence conditions. Internal labels are preserved for
reproducibility.}\label{tab:conditions}
\resizebox{\textwidth}{!}{%
\begin{tabular}{p{2.8cm}p{3.8cm}p{5.0cm}r}
\toprule
Display Label & Internal Label & Description & N\\
\midrule
C1: Chinese-only correct & \texttt{C1\_ZH\_ONLY\_CORRECT} &
Correct evidence in Chinese. & 50\\
C2: English-only correct & \texttt{C2\_EN\_ONLY\_CORRECT} &
Correct evidence in English. & 50\\
C3: Bilingual consistent & \texttt{C3\_BILINGUAL\_CONSISTENT} &
Both languages support same candidate. & 50\\
C4: Chinese true, EN false & \texttt{C4\_ZH\_TRUE\_EN\_FALSE} &
Chinese supports reference; English supports competing candidate. & 50\\
C5: English true, ZH false & \texttt{C5\_EN\_TRUE\_ZH\_FALSE} &
English supports reference; Chinese supports competing candidate. & 50\\
C6: Conflict with optional noise & \texttt{C6\_CONFLICT\_WITH\_NOISE} &
Bilingual conflict; noise present in 34 of 50 cases. & 50\\
\bottomrule
\end{tabular}%
}
\end{table}

\subsection{Construction and Leakage Control}

\begin{figure*}[t]
    \centering
    \includegraphics[width=0.95\textwidth]{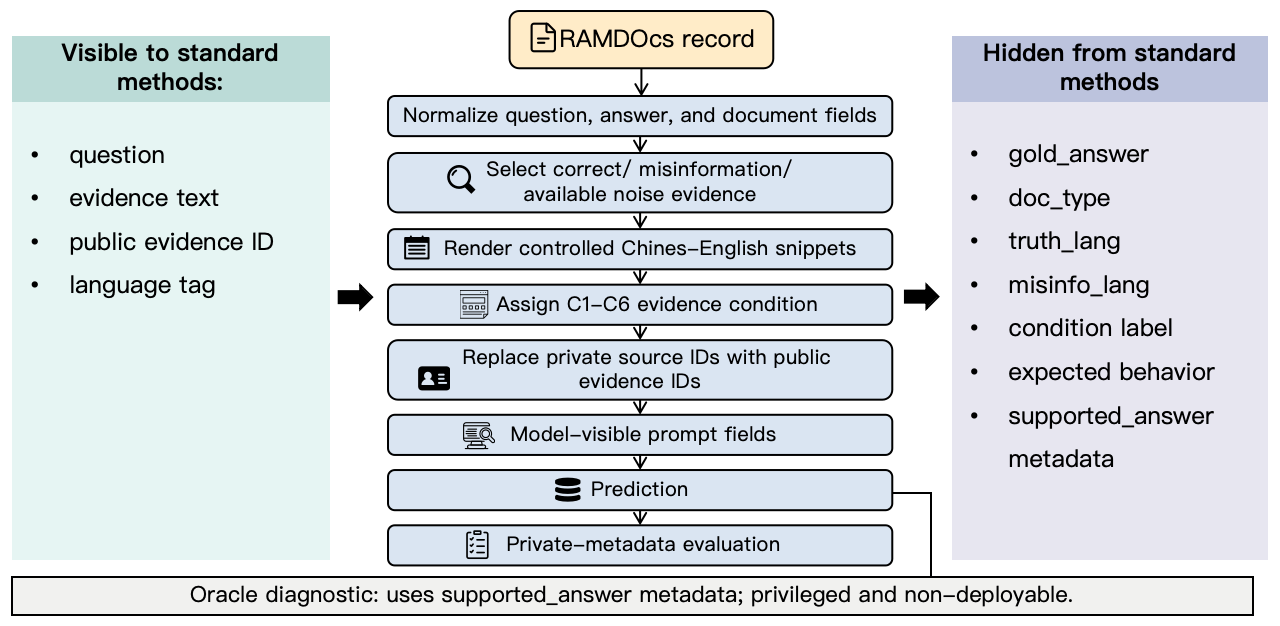}
    \caption{Construction and leakage-control flow of X-RAMDocs-ZHEN. Standard
    methods receive only the question, visible evidence text, public evidence
    IDs, and language tags. Gold answers, document roles, conflict labels,
    language roles, expected behavior, and supported-answer metadata are hidden
    from standard prompts and used only for construction and automatic
    evaluation. The oracle diagnostic is the only method that accesses
    document-level supported-answer metadata and is therefore non-deployable.}
    \label{fig:dataset-flow}
\end{figure*}

Figure~\ref{fig:dataset-flow} summarizes the construction and leakage-control
design of X-RAMDocs-ZHEN. The benchmark starts from RAMDocs records, normalizes
question, answer, and document fields, renders controlled Chinese--English
evidence snippets, assigns one of the six balanced evidence conditions, and
replaces private source identifiers with public evidence IDs. Standard methods
receive only public prompt fields: the question, visible evidence text, public
evidence IDs, and language tags. Gold answers, document roles, language roles,
condition labels, expected behavior, and supported-answer metadata are hidden
from standard prompts and used only for construction and automatic evaluation.

This separation ensures that standard models cannot infer the target behavior
from hidden labels or construction metadata. The oracle diagnostic is the only
exception: it accesses document-level supported-answer metadata and is therefore
treated only as a privileged upper-bound analysis, not as a deployable baseline.

\subsection{Naturalized Stress Test}

Because the original benchmark uses controlled evidence templates, we construct a
naturalized stress test. This deterministic transformation removes explicit
answer templates while preserving answer candidate strings and private metadata,
testing whether methods depend on surface cues such as ``the answer is'' or
``\begin{CJK}{UTF8}{gbsn}答案是\end{CJK}''. The full transformed set covers
all 300 samples and 534 documents with zero remaining templates and perfect
candidate and metadata preservation. We evaluate a 100-sample approximately
balanced subset; it remains controlled prose rather than natural retrieval.

\section{Methods}

\subsection{Baselines and Diagnostics}

The \textbf{naive RAG} baseline concatenates all visible evidence into one
prompt and asks the model to return an answer, evidence identifiers, a conflict
flag, and notes. The \textbf{evidence-normalized RAG} baseline groups visible
evidence by language before a single model call, testing whether explicit
language organization helps. The \textbf{rule-only extractor} applies
conservative Chinese and English surface patterns, groups normalized candidates
deterministically, and generates a fixed response with zero model calls; it
measures template dependence, not semantic extraction. The \textbf{oracle
diagnostic} replaces document-level extraction with private
\texttt{supported\_answer} metadata and applies the same downstream grouping
and aggregation; it is privileged and non-deployable, used only to estimate the
impact of imperfect extraction.

\subsection{X-MADAM-RAG}

\begin{figure*}[t]
    \centering
    \includegraphics[width=0.95\textwidth]{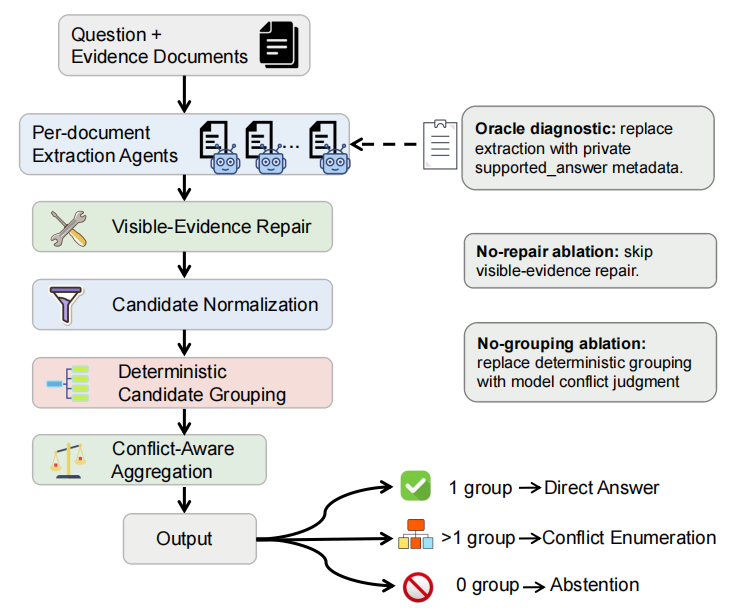}
    \caption{X-MADAM-RAG pipeline. Each evidence document is processed
    independently by an extraction agent. Missing or unusable candidates may be
    repaired from visible evidence only. Normalized candidates are then grouped
    deterministically. A single group yields a direct answer, multiple groups
    yield conflict enumeration, and no group yields abstention. The no-repair
    and no-grouping ablations remove the corresponding components, while the
    oracle diagnostic replaces extraction with privileged document-level
    metadata.}
    \label{fig:xmadam-pipeline}
\end{figure*}

Figure~\ref{fig:xmadam-pipeline} gives an overview of X-MADAM-RAG. The pipeline
decomposes evidence-conflict handling into document-level extraction,
visible-evidence repair, candidate normalization, deterministic grouping, and
conflict-aware aggregation. Each evidence document is processed independently by
an extraction agent that receives only the question, public evidence ID, language
tag, and visible evidence text.

When extraction fails or returns an unusable candidate, a label-blind repair
module searches parsed fields, raw agent output, and visible evidence; it cannot
access hidden metadata. Repaired candidates are normalized and grouped
deterministically. A single candidate group yields a direct answer, multiple
groups yield explicit conflict enumeration, and no candidate yields abstention.

We evaluate two ablations. The \textbf{no-repair} variant removes the repair
stage while keeping extraction, grouping, and aggregation. The
\textbf{no-grouping} variant replaces deterministic grouping with a model-based
conflict decision. These isolate the contribution of each component. The oracle
diagnostic replaces only the extraction stage with private document-level
supported-answer metadata, distinguishing extraction failures from downstream
grouping and aggregation failures.

\clearpage
\section{Experimental Setup}

All LLM experiments use local Qwen2.5-7B-Instruct inference with deterministic
decoding (temperature~0), bfloat16 weights, SDPA attention, and a maximum of
512 new tokens.\footnote{Python 3.10.20, PyTorch 2.6.0+cu124, Transformers
4.57.6, NVIDIA RTX 4090 ($\approx$48\,GiB VRAM).} Only one model is evaluated.
We evaluate on the original 300-sample benchmark and the 100-sample naturalized
stress subset; component ablations are run on the original benchmark only.

We use five automatic metrics. \textbf{Strict accuracy} measures normalized
lexical match between the returned answer and the reference. \textbf{Conflict-aware
success} requires conflict recognition and reference-candidate preservation on
conflict samples, and reduces to strict correctness on non-conflict samples.
\textbf{Conflict F1} evaluates expected versus predicted conflict.
\textbf{Conflict awareness} measures explicit conflict expression among
expected-conflict cases. \textbf{Unsupported-candidate rate} measures whether
generated candidates fail to lexically match any evidence-supported candidate in
private metadata. None of these metrics constitute semantic entailment, factual
verification, or human judgments.

For original-benchmark comparisons we use 1,000 bootstrap resamples with
seed~13; paired sign tests are exploratory and uncorrected for multiple
comparisons.

\section{Results and Analysis}

\subsection{Original Controlled Benchmark}

\begin{table}[t]
\caption{Results on the original controlled benchmark (300 samples). Best
values per column in bold. Rule-only and oracle rows are diagnostic: rule-only
measures template dependence; oracle uses privileged \texttt{supported\_answer}
metadata and is non-deployable.}\label{tab:main}
\begin{tabular}{lrrrrrr}
\toprule
Method & Coverage & Strict & Conf.-aware & Conf.\ F1 & Awareness &
Unsupported\\
\midrule
Naive & 1.0000 & 0.8700 & 0.8667 & 0.9579 & 0.7400 & 0.0800\\
Evidence-normalized & 1.0000 & 0.8900 & 0.8867 & 0.9766 & 0.7533 & 0.1067\\
Rule-only extractor & 1.0000 & \textbf{1.0000} & \textbf{1.0000} &
\textbf{1.0000} & \textbf{1.0000} & \textbf{0.0000}\\
X-MADAM-RAG & 1.0000 & 0.9667 & 0.9767 & 0.9804 & \textbf{1.0000} & 0.0200\\
Oracle extraction & 1.0000 & \textbf{1.0000} & \textbf{1.0000} &
\textbf{1.0000} & \textbf{1.0000} & \textbf{0.0000}\\
\bottomrule
\end{tabular}
\end{table}

Among deployable LLM methods, X-MADAM-RAG performs best on the original
benchmark. It improves strict accuracy over evidence-normalized RAG by 0.0767
and lowers unsupported-candidate rate from 0.1067 to 0.0200 (95\% bootstrap
interval for the strict-accuracy gap: $[0.0367, 0.1167]$; exploratory sign-test
$p=0.0004$). However, the rule-only extractor also achieves perfect performance,
showing that the original benchmark is strongly template-regular. Original
benchmark results should therefore be interpreted as controlled diagnostic
performance rather than evidence of semantic extraction robustness.

\subsection{Component Ablations}

\begin{table}[t]
\caption{Component ablations on the original benchmark.}\label{tab:ablation}
\small
\begin{tabular}{p{3.2cm}rrrrrr}
\toprule
Variant & Coverage & Strict & Conf.-aware & Conf.\ F1 & Awareness &
Unsupported\\
\midrule
X-MADAM without repair & 1.0000 & 0.8033 & 0.6967 & 0.7089 & 0.6467 & 0.0933\\
X-MADAM without det.\ grouping & 1.0000 & 0.8600 & 0.9433 & 0.9521 & 0.9933 &
0.0367\\
Full X-MADAM & 1.0000 & \textbf{0.9667} & \textbf{0.9767} & \textbf{0.9804} &
\textbf{1.0000} & \textbf{0.0200}\\
\bottomrule
\end{tabular}
\end{table}

Both components contribute within the original X-MADAM pipeline. Disabling
repair produces the largest drop, reducing strict accuracy from 0.9667 to
0.8033. Replacing deterministic grouping with model conflict judgment also
degrades performance, especially strict accuracy and unsupported-candidate rate.

\subsection{Naturalized Stress Test}

\begin{table}[t]
\caption{Results on the 100-sample naturalized stress
subset.}\label{tab:stress}
\begin{tabular}{lrrrrrr}
\toprule
Method & Coverage & Strict & Conf.-aware & Conf.\ F1 & Awareness &
Unsupported\\
\midrule
Rule-only extractor & 1.0000 & 0.0000 & 0.0000 & 0.0000 & 0.0000 & 0.0000\\
Naive & 1.0000 & 0.6500 & 0.6300 & 0.8762 & 0.6200 & 0.1900\\
Evidence-normalized & 1.0000 & \textbf{0.6800} & \textbf{0.6600} & 0.9495 &
\textbf{0.6400} & 0.3000\\
X-MADAM-RAG & 1.0000 & 0.3000 & 0.2600 & 0.1481 & 0.5800 & 0.0500\\
Oracle extraction & 1.0000 & \textbf{1.0000} & \textbf{1.0000} &
\textbf{1.0000} & \textbf{1.0000} & \textbf{0.0000}\\
\bottomrule
\end{tabular}
\end{table}

\begin{figure}[t]
    \centering
    \includegraphics[width=\columnwidth]{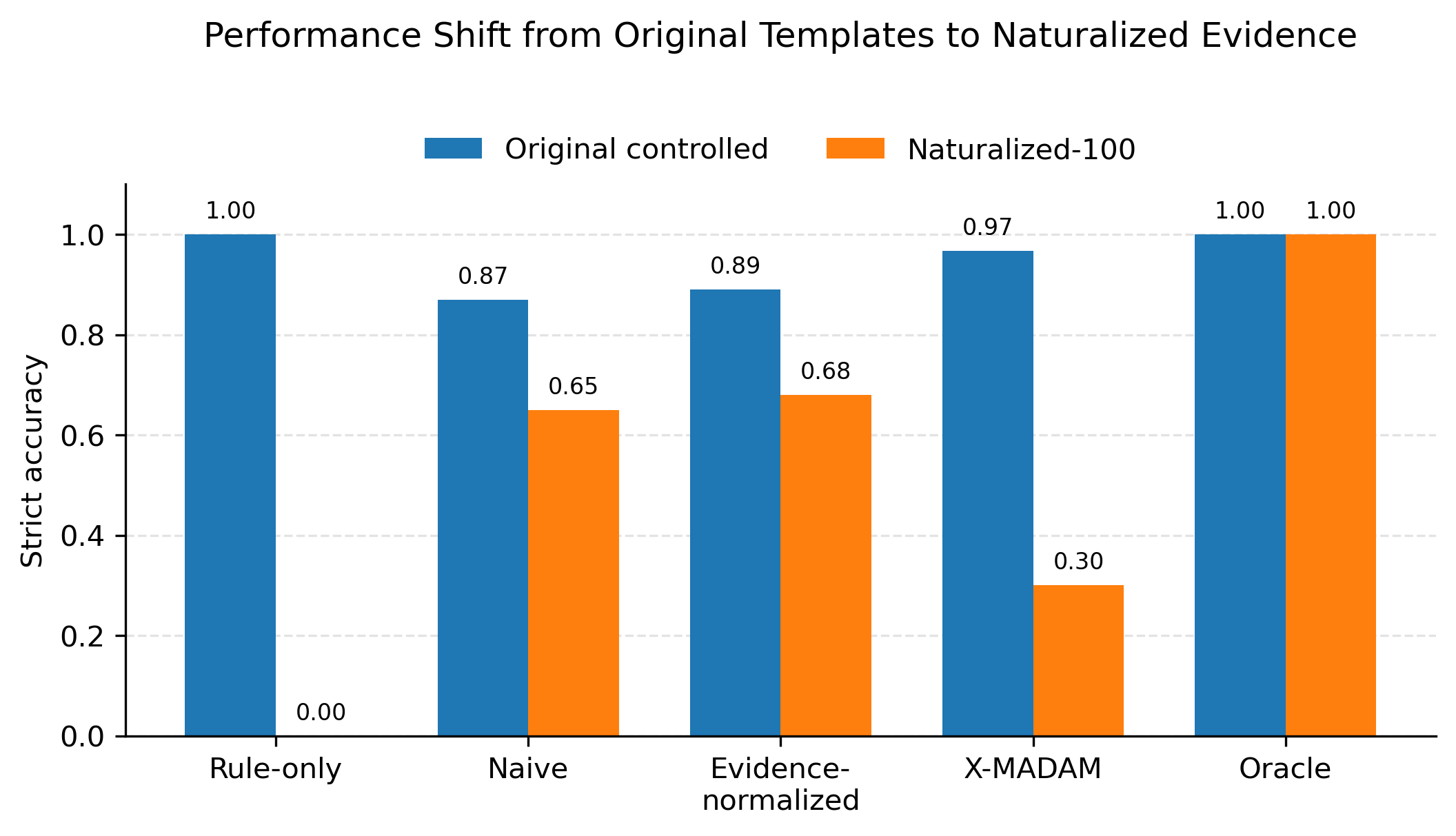}
    \caption{Strict accuracy before and after naturalization. Rule-only
    extraction drops from 1.0000 to 0.0000 after explicit answer templates are
    removed. X-MADAM-RAG also drops from 0.9667 to 0.3000, while the oracle
    remains at 1.0000, indicating that document-level extraction is the main
    bottleneck under naturalized phrasing.}
    \label{fig:original-vs-naturalized}
\end{figure}

Table~\ref{tab:stress} reports results on the naturalized-100 stress subset,
and Figure~\ref{fig:original-vs-naturalized} compares strict accuracy before and
after naturalization. Naturalization removes the explicit rule cues: the rule-only
extractor drops from 1.0000 strict accuracy on the original benchmark to 0.0000
on the naturalized subset. This confirms that its original success depends on
surface templates rather than semantic extraction.

X-MADAM-RAG also degrades substantially, from 0.9667 strict accuracy on the
original benchmark to 0.3000 on the naturalized subset. By contrast, the naive
and evidence-normalized single-call baselines retain 0.6500 and 0.6800 strict
accuracy, respectively. This shows that the current X-MADAM extractor is
sensitive to naturalized phrasing and should not be described as robust in this
setting.

The oracle diagnostic remains at 1.0000 in both settings. Since the oracle
replaces only document-level extraction with privileged supported-answer metadata
while retaining the downstream grouping and aggregation stages, this pattern
identifies extraction as the principal bottleneck. The naturalized stress test is
therefore a negative robustness result, but also a useful diagnostic result: it
shows that improving document-level extraction is more urgent than adding further
aggregation complexity.

Additional breakdowns by answer type, conflict direction, and C6 optional-noise
subgroups are omitted from the main paper for space; they are consistent with
the main findings.

\section{Discussion and Limitations}

On the original benchmark, X-MADAM-RAG is the strongest deployable LLM method
and its internal components are demonstrably useful: repair recovers missing
candidates from visible evidence and deterministic grouping improves auditability
and performance.

At the same time, the perfect rule-only result shows that the original benchmark
is highly template-regular, limiting what can be concluded from original
benchmark performance. X-RAMDocs-ZHEN v0.1 is best understood as a controlled
diagnostic tool for evidence-conflict behavior, not a general robustness
benchmark.

The naturalized stress test sharpens this conclusion. X-MADAM-RAG degrades below
single-call baselines while the oracle remains perfect, indicating that the
extraction stage---not grouping or aggregation---is the central bottleneck.
Conflict-aware aggregation alone is insufficient: a system must first extract
candidate answers robustly from varied evidence forms. Future work should focus
on document-level extraction under paraphrase, richer answer normalization, and
semantic candidate verification.

This study has several limitations. X-RAMDocs-ZHEN uses controlled snippets
derived from RAMDocs fields rather than naturally retrieved multilingual passages.
The original benchmark is template-regular and cannot establish semantic
extraction ability. The naturalized stress test is still deterministic controlled
prose. All LLM experiments use one model, Qwen2.5-7B-Instruct; results may
differ for larger or closed models. The benchmark covers only Chinese and
English. All metrics are automatic lexical diagnostics, not human semantic
judgments. No completed human audit is available. The oracle is privileged and
non-deployable. Our claims are limited to controlled Chinese-English
evidence-conflict diagnosis; we do not claim broad hallucination detection,
robustness to natural retrieval, or human-validated faithfulness.

\section{Conclusion}

We presented X-RAMDocs-ZHEN, a controlled leakage-aware Chinese-English
benchmark for diagnosing evidence conflict in RAG, and studied X-MADAM-RAG, an
interpretable pipeline based on per-document extraction, visible-evidence repair,
deterministic candidate grouping, and conflict-aware aggregation. On the original
controlled benchmark, X-MADAM-RAG outperforms single-call LLM baselines and
ablations confirm that repair and deterministic grouping contribute. However, a
rule-only extractor achieves perfect performance, revealing strong template
regularity. A naturalized stress test removes explicit rule cues: X-MADAM-RAG
degrades below single-call baselines while a privileged oracle remains perfect,
identifying extraction robustness as the central unresolved problem. Future work
should evaluate natural retrieval, complete human audit, test additional models
and languages, and develop more robust document-level extraction and semantic
candidate verification.

\begin{credits}
\subsubsection{\ackname} The authors thank the contributors of the RAMDocs
dataset, on which X-RAMDocs-ZHEN is based.

\subsubsection{\discintname}
The authors have no competing interests to declare that are relevant to the
content of this article.
\end{credits}
\bibliographystyle{splncs04}
\bibliography{references}
\end{document}